\documentclass{article}

\PassOptionsToPackage{numbers,compress}{natbib}

\usepackage[preprint]{neurips_2023}
\usepackage{subfig}
\usepackage[ruled,vlined,linesnumbered]{algorithm2e}
\usepackage{algorithmic}

\usepackage[utf8]{inputenc} 
\usepackage[T1]{fontenc}    
\usepackage{hyperref}       
\usepackage{url}            
\usepackage{booktabs}       
\usepackage{amsfonts}       
\usepackage{nicefrac}       
\usepackage{microtype}      
\usepackage{xcolor}         
\usepackage{amsmath}
\usepackage{makecell}
\usepackage{multirow}
\usepackage{graphicx}

\title{Exploring Data Redundancy in Real-world Image Classification through Data Selection}

\author{%
    \textbf{
    Zhenyu Tang$^{\,1,2}$
    $\quad$
    Shaoting Zhang$^2$ 
    $\quad$
    Xiaosong Wang$^2$
    }
    \\\\
    $^1$Shanghai Jiao Tong University$\quad$
    $^2$Shanghai AI Laboratory\\
}

\date{}

\begin{document}

\maketitle

\begin{abstract}
    Deep learning models often require large amounts of data for training, leading to increased costs. It is particularly challenging in medical imaging, i.e., gathering distributed data for centralized training, and meanwhile, obtaining quality labels remains a tedious job. Many methods have been proposed to address this issue in various training paradigms, e.g., continual learning, active learning, and federated learning, which indeed demonstrate certain forms of the data valuation process. However, existing methods are either overly intuitive or limited to common clean/toy datasets in the experiments. In this work, we present two data valuation metrics based on Synaptic Intelligence and gradient norms, respectively, to study the redundancy in real-world image data. Novel online and offline data selection algorithms are then proposed via clustering and grouping based on the examined data values. Our online approach effectively evaluates data utilizing layerwise model parameter updates and gradients in each epoch and can accelerate model training with fewer epochs and a subset (e.g., 19\%-59\%) of data while maintaining equivalent levels of accuracy in a variety of datasets. It also extends to the offline coreset construction, producing subsets of only 18\%-30\% of the original. The codes for the proposed adaptive data selection and coreset computation are available (\url{https://github.com/ZhenyuTANG2023/data_selection}).
\end{abstract}

\section{Introduction}

Recent progress of large language models and foundation models\cite{brown2020language,fedus2022switch,he2022masked,lin2021m6,radford2021learning,touvron2023llama,yu2022coca,yuan2021florence} demonstrates the amazing generalization ability of deep learning models, trained on large amounts of data and prompted (with few-shot samples) for downstream applications. However, the demand for data increases exponentially as the test error decreases\cite{kaplan2020scaling, sorscher2022beyond}. 
In medical imaging, models are data-hungry but suffer from a lack of high-quality labels. Moreover, normal samples often take up the majority of medical image datasets, potentially resulting in data redundancy for training. While obtaining enough amount of data is crucial for model training, the question remains whether such a large amount of data is truly necessary\cite{birodkar2019semantic,paul2021deep,vodrahalli2018all}. Furthermore, different stages of training may require different data for precise and efficient training. 
The problem can be approached in two steps: 1) investigating how each data sample will influence model training and 2) utilizing this information to facilitate data selection and then more efficient model training.

The valuation and selection of data have been explored in several research fields. In federated learning studies, the concept of data valuation is discussed\cite{ghorbani2020distributional,shyn2021fedccea,wang2020principled,wei2020efficient} in the context of incentive mechanisms and fairness guarantees. In addition to using evaluation as an incentive for attracting more clients and data, other research focuses on data evaluation for selection purposes. In active learning, the data without labels are first evaluated using the trained models, and the most valuable samples are picked for annotation by oracles and added to the model training. Alternatively, data pruning and coreset selection methods evaluate the representativeness of subsets within a dataset and minimize the original dataset into a core set, hoping the data reduction will not harm the training performance. Furthermore, replay-based continual learning methods select a dominant subset for tackling the catastrophic forgetting issue when new data are included in the training. 
acks (this paragraph is not well written)

\begin{figure}[t]
    \centering
    \includegraphics[width=0.8\textwidth]{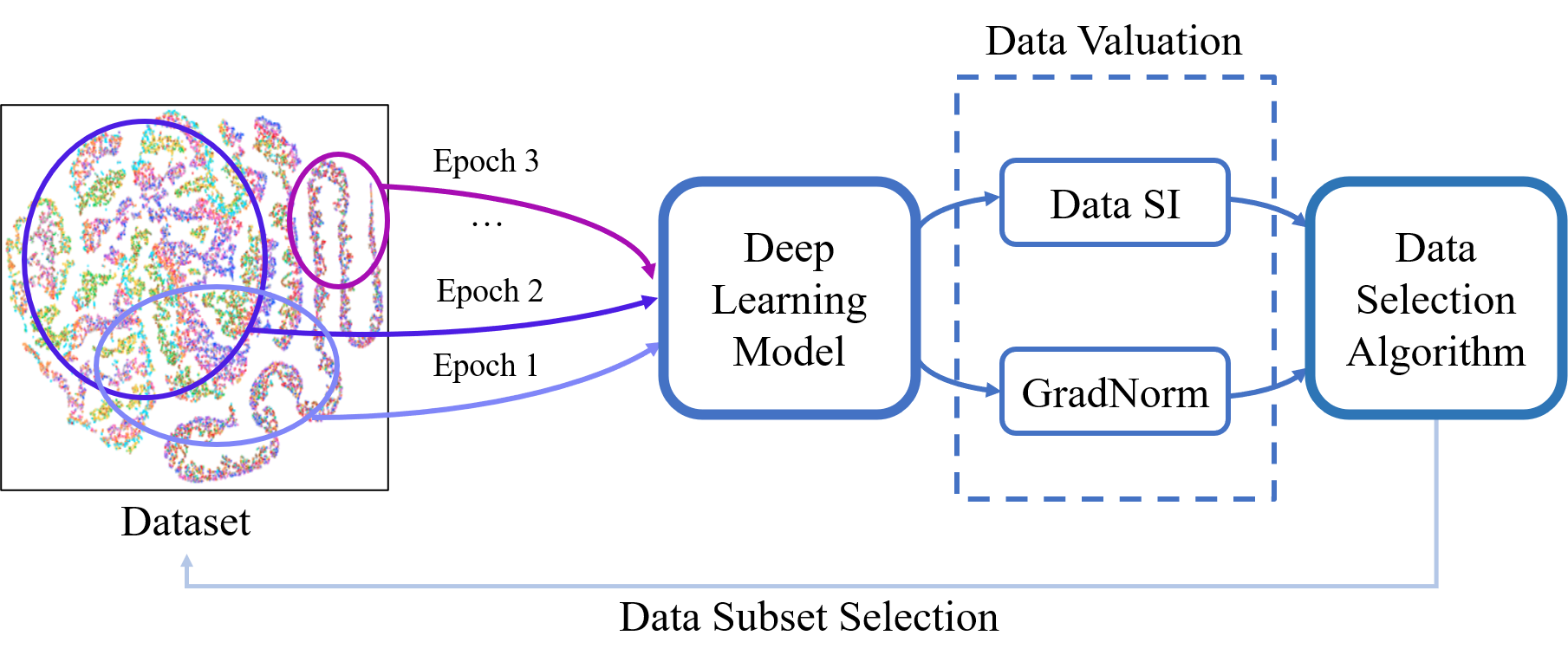}
    \caption{Overview of our data valuation and selection framework. We evaluate data on the given deep learning model by two proposed metrics, Data SI and GradNorm, and \textit{adaptively} select a \textit{varying} number of data samples for model training in each epoch.}
    \label{fig:overview}
\end{figure}

In this paper, we propose to adaptively select a subset of data with varying sizes using layerwise information of the network at each epoch of the training process (shown in Fig.~\ref{fig:overview}), which is experimented with on a variety of real-world image data. The presented work is most close to recent gradient-based data selection works~\cite{paul2021deep,killamsetty2021grad}, which select a fixed number of samples, use gradients from the late prediction output, and are often evaluated on toy datasets. Our contributions in data valuation and selection are 3-fold:

\begin{enumerate}
    \item Presenting two gradient-based data valuation metrics by considering the layerwise information of the network
    and utilizing them to investigate the redundancy in real-world images;
    \item Proposing online and offline data selection algorithms based on the gradient-based metrics and further demonstrate techniques for efficient and precise training;
    \item Conducting experiments to justify the superior performance of our online adaptive data selection and coreset selection algorithms, using a variety of real-world datasets as well as on different network architectures.
\end{enumerate}

\section{Related Work}

\textbf{Active Learning and Coreset.} Active learning involves selecting unlabeled data and sending it to oracles for annotation. Current active learning methods include uncertainty-based\cite{citovsky2021batch,netzer2011reading,nguyen2022measure,roth2006margin,wang2014new,zhan2022comparative}, density-based and hybrid strategies. Yet active learning primarily reduces the labeling cost as all annotated samples used in training accumulate over time. Coreset\cite{mirzasoleiman2020coresets,sener2017active} and data pruning\cite{chitta2021training,mindermann2022prioritized,paul2021deep,sorscher2022beyond} approaches select a subset of data for a full model update. Adaptive data subset selection\cite{killamsetty2021grad} selects dynamic subsets every few epochs. However, in each iteration, the selected data size is fixed and preset. Moreover, many data pruning approaches are limited to common clean/toy datasets in the experiments\cite{paul2021deep,killamsetty2021grad} (e.g., CIFAR-10). Given that warm-start full-set training epochs can significantly improve the performance of adaptive selection methods, we introduce selection methods with varying subset sizes in this paper and evaluate them on a variety of cross-domain datasets.

\textbf{Replay-based Continual Learning.} Replay-based continual learning\cite{chaudhry2019tiny,isele2018selective,rebuffi2017icarl,rolnick2019experience} tackles the issue of catastrophic forgetting by saving data from previous tasks in a memory buffer and utilizing them to reinforce knowledge during training subsequent tasks \cite{merlin2022practical}. Due to the limited memory size, the selection and discarding procedures inherently involve cross-task data valuation. In practice, stream-based uniform random sampling, a.k.a. reservoir sampling, is often the most effective strategy\cite{chaudhry2019tiny,isele2018selective}. This paper aims at in-task non-uniform data valuation and may provide inspiration for cross-task valuation methods applicable to replay-based continual learning.

\textbf{Fairness and Incentive Mechanisms for Federated Learning.} Ensuring fairness guarantees and implementing effective incentive mechanisms are essential components of federated learning systems, as they serve to increase participation by attracting a larger number of clients. The concept of data valuation is frequently employed in the literature in these fields as a means of assessing the value of the data held by individual clients. Common approaches to data valuation include evaluations based on the volume of data\cite{jiao2020toward,nishio2020estimation,zeng2020fmore,zhan2020learning} and historical data\cite{lyu2020collaborative,lyu2020towards,zhang2020hierarchically,zhang2021incentive}. While these approaches are relatively straightforward to implement, they may be considered somewhat simplistic. Some other approaches based on game theory, primarily Shapley's Value\cite{ghorbani2020distributional,liu2022gtg,ma2021transparent,nagalapatti2021game,song2019profit,wang2019interpret,wang2020principled,wei2020efficient}, are computationally intensive and often approximated using historical gradients. This paper focuses on the non-trivial and computationally efficient data valuation metrics, which can potentially be applied to federated learning systems\cite{jiang2023fair}. A brief comparison between the related work and our data valuation and selection can be seen in Table~\ref{tab:related-work-compare}.

\begin{table}[tb]
    \centering
    \caption{Brief comparison between related work and our work. Our method uses adaptive size of data subset for model training and depends on model gradients and activations.}
    \begin{tabular}{ccccc}
    \toprule
         & Selection & Training Data & Subset Size & Dependency \\
    \midrule
        Active Learning & Adaptive & Annotated set & Fixed & Data, Activations\\
        Data Pruning & Fixed & Subset & Fixed & Data, Labels, Gradients\\
        Replay-based CL & Fixed & Subset and new dataset & Fixed & Data, Labels\\
        Fairness in FL & - & Private datasets & - & Gradients\\
    \midrule
        Ours & Adaptive & Subset & Adaptive & Gradients, Activations\\
    \bottomrule
    \end{tabular}
    \label{tab:related-work-compare}
\end{table}

\section{Method}

\subsection{Data Valuation via Synaptic Intelligence}

For defying catastrophic forgetting in continual learning, \cite{zenke2017continual} develops Synpatic Intelligence (SI) which constrains the shift of important parameters from previous tasks when learning a new task through regularization. SI evaluates the importance of the parameters with 
the allocation of the loss change path integrals to the parameters, as shown in Eq.~\ref{eq:synaptic-intelligence}, where $\mathcal L$ is the loss, $\pmb\theta=[\theta_1,\theta_2,\dots,\theta_n]^T$ is a vector of $n$ model parameters, and $\mathcal X=\{\pmb x_1, \pmb x_2, \dots, \pmb x_m\}$ represents the set of $m$ data samples used in training. The integrals are estimated using first-order Taylor expansion, as shown in Eq.~\ref{eq:def-si}. The denominator for carrying the same unit as in the original paper is not included here since SI is not used for regularization.

\begin{equation}
\begin{aligned}
    \mathcal L(\pmb\theta+\Delta \pmb\theta, \mathcal X)-\mathcal L(\pmb\theta, \mathcal X) &= \int_{\pmb\theta}^{\pmb\theta+\Delta\pmb\theta} \left(\dfrac{\partial \mathcal L}{\partial \pmb\theta}\right)^T {\mathrm d}\pmb\theta   =\sum_{i=1}^{n}\int_{\theta_i}^{\theta_i+\Delta\theta_i} \dfrac{\partial\mathcal L}{\partial \theta_i}{\mathrm d}\theta_i
\end{aligned}
\label{eq:synaptic-intelligence}
\end{equation}

\begin{equation}
    \label{eq:def-si}
    SI(\theta_i, \mathcal X)\stackrel{\text{def}}{=}-\int_{\theta_i}^{\theta_i+\Delta\theta_i}\dfrac{\partial\mathcal L(\pmb\theta, \mathcal X)}{\partial \theta_i}{\mathrm d}\theta_i\approx -\dfrac{\partial\mathcal L(\pmb\theta, \mathcal X)}{\partial \theta_i}\Delta\theta_i 
\end{equation}

SI is also used to analyze the contribution of layer-wise parameters in \cite{lan2019lca}. Inspired by SI, we further allocate the loss change on data, as shown in Eq.~\ref{eq:our-method}. 

\begin{equation}
    \label{eq:our-method}
    \begin{aligned}
    \mathcal L(\pmb\theta+\Delta \pmb\theta, \mathcal X_{})-\mathcal L(\pmb\theta, \mathcal X_{}) &=\dfrac{1}{m}\sum_{j=1}^m \sum_{i=1}^n \int_{\theta_i}^{\theta_i+\Delta\theta_i}\dfrac{\partial\mathcal L(\pmb\theta, \{\pmb x_j\})}{\partial \theta_i}{\mathrm d}\theta_i
\end{aligned}
\end{equation}

Data SI is defined as the loss change allocated on a subset of data approximated by first order Taylor expansion, where $\Delta\theta_i$ can be estimated ahead using gradient accumulation on the selected set given a certain optimizer, in Eq.~\ref{eq:data-si-def}. The parameters can be further divided into $s$ non-overlapping subsets $\pmb \theta_1, \pmb \theta_2, \dots, \pmb \theta_s$, 
e.g., $\pmb\theta_i$ could be a set of parameters from a layer in the network.

\begin{equation}
    \label{eq:data-si-def}
    DataSI(\pmb \theta, \mathcal X)\stackrel{\text{def}}{=}-\dfrac{1}{m}\sum_{j=1}^m \sum_{i=1}^n \dfrac{\partial\mathcal L(\pmb\theta, \{\pmb x_j\})}{\partial \theta_i}\Delta \theta_i=\sum_{i=1}^s DataSI(\pmb\theta_i, \mathcal X)
\end{equation}

In practice, we can compute the parameter updates $\{\Delta\theta_i\}$ on the whole training set. However, the final Data SIs computed on the selected subsets may vary significantly from the estimation. To mitigate such estimation error, the target is set to greedily maximizing loss change, and vanilla SGD (vSDG) is taken as an example. The Data SI of the whole training set is transformed to the norm of averaged gradient vectors as in Eq.~\ref{eq:datasi-to-gradnorm}, where $\gamma$ is the learning rate and $G_{i,j}=\dfrac{\partial \mathcal L(\pmb\theta, \{\pmb x_j\})}{\partial \theta_i}$. 

\begin{equation}
    \label{eq:datasi-to-gradnorm}
    DataSI_{vSGD}(\pmb \theta, \mathcal X)=\dfrac{\gamma}{m^2} \mathbf 1^T\mathbf G^T\mathbf G\mathbf 1=\gamma\left|\left|\mathbf G\dfrac{\mathbf 1}{m}\right|\right|^2
\end{equation}

For a subset of the training set, the uniform weighting vector $\mathbf 1/m$ could be replaced with a certain selection weighting vector $\mathbf w$. The data samples with maximum gradient vector norms should be selected given the convexity of the target function in Eq.~\ref{eq:gradnorm-deriv}. Therefore, the gradient norm can be taken as another evaluation metric for data valuation, defined as GradNorm (Eq.~\ref{eq:gradnorm-def}). 

\begin{equation}
    \label{eq:gradnorm-deriv}
    \arg\max\limits_{\mathbf w} \gamma \mathbf w^T\mathbf G^T\mathbf G\mathbf w, {\rm \;s.t.\;} \mathbf w^T\mathbf 1=1, w_j\ge 0 \; \forall j\in \{1,2,\dots, m\}
\end{equation}

\begin{equation}
    \label{eq:gradnorm-def}
    GradNorm(\pmb\theta, \mathcal X)\stackrel{\text{def}}{=}\dfrac{1}{m}\sum_{j=1}^m\left|\left|\dfrac{\partial \mathcal L(\pmb\theta, \{\pmb x_j\})}{\partial \pmb\theta}\right|\right|^2
\end{equation}

From the perspective of greedily maximizing the loss change of the selected subset, we reach the same conclusion as \cite{paul2021deep}. According to their finding, the marginal loss change of removing a sample is bounded by a specific instance of our defined GradNorm (approximated directly with the difference between the prediction and the one-hot label).

\begin{figure}[tb]
    \centering
    \includegraphics[width=0.8\textwidth]{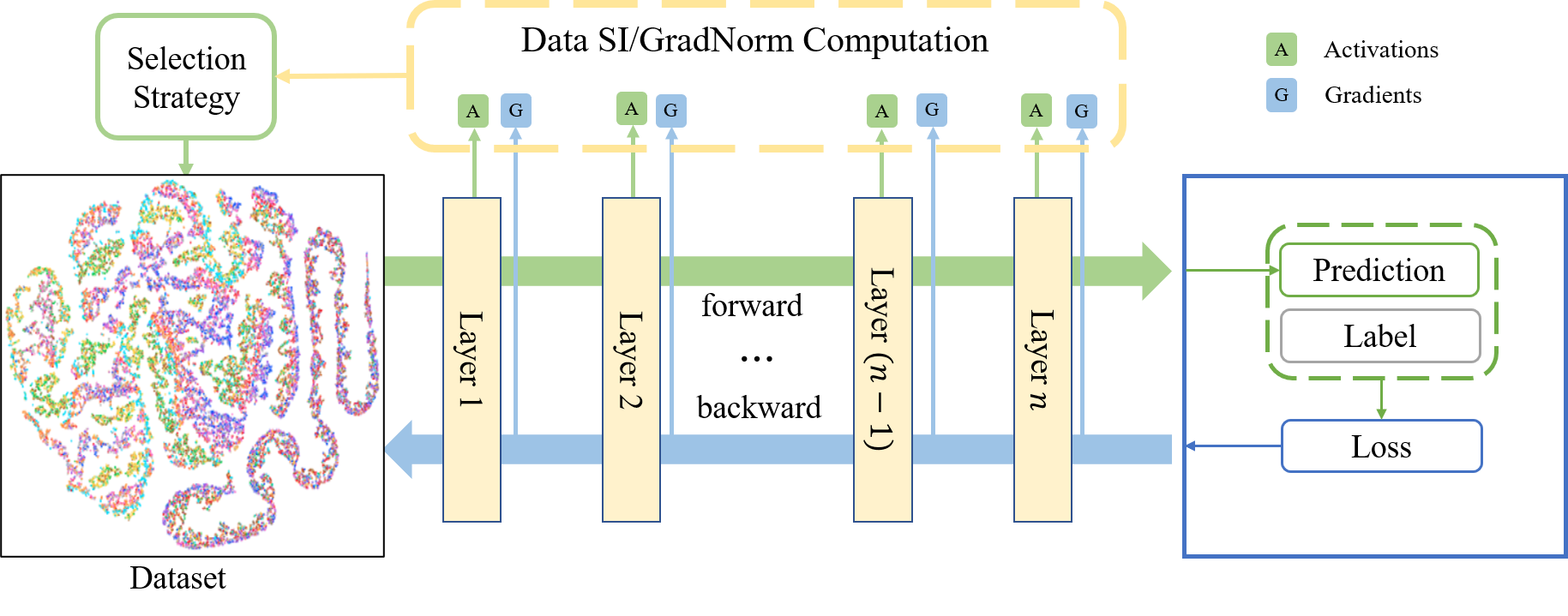}
    \caption{Adaptive Gradient-based Data Selection Pipeline.}
    \label{fig:pipeline}
\end{figure}

\subsection{Sample Clustering and Grouping via Data Valuation}

Fig.~\ref{fig:pipeline} illustrates the proposed data selection pipeline. For each data sample, the Data SIs could be computed for each model parameter as a feature vector (with a length equal to the total number of parameters) for the effectiveness measure in model training. 
In practice, we divide the model parameters layerwise (averaged Data SI for each layer) and construct a feature vector of Data SI (with a length equal to the number of layers) for each sample. To select data samples that contribute to certain layers, the data samples are then clustered using the computed feature via K-Means parameterized by $k$. There is often a large cluster centered around the axis origin in trained models observed in the experiment, and hence when a cluster occupies over $\tau=50\%$ of the training data, it will be considered less valuable for further training and therefore discarded in the next epoch. Algorithm~\ref{alg:data-si-sel} demonstrates our sample clustering method.

\begin{algorithm}[t]
    \caption{Clustering Samples using Data SI}
    \label{alg:data-si-sel}
    \begin{algorithmic}[1]
    \REQUIRE Training set $\mathcal X$, the number of epochs $n$, the number of clusters $k$, cluster size threshold $\tau$, learning rate $\gamma$, learning rate scheduler $s$
    \FOR{epoch$\in\{1,2,\dots, n\}$}
        \STATE For all $\pmb x_j$ in $\mathcal X$, compute $DataSI(\pmb\theta, \{\pmb x_j\})$
        \STATE K-Means cluster samples in $\mathcal X$ using $DataSI(\pmb\theta, \{\pmb x_j\})$ as the feature for $\pmb x_j$
        \STATE $\alpha\leftarrow$ $\dfrac{\rm the\; size\; of\; the\; largest\; cluster}{|\mathcal X|}$
        \IF{$\alpha > \tau$}
            \STATE $\gamma\leftarrow (1-\alpha)\gamma$
            \STATE Train the model on $\mathcal X$ except for the largest cluster
            \STATE $\gamma\leftarrow \dfrac{\gamma}{(1-\alpha)}$
            \STATE $\gamma\leftarrow s(\gamma)$
        \ELSE
            \STATE Train the model on $\mathcal X$
        \ENDIF
    \ENDFOR
    \end{algorithmic}
\end{algorithm}

Regarding grouping samples using GradNorm, we find that including samples with large outlier GradNorm sometimes results in underperforming models in the experiment. A small GradNorm value indicates that a sample contributes little to the training process, while samples with large GradNorm may imply noisy or hard samples for the current model. We set the upper and lower thresholds for GradNorms and select the data samples whose values are within the thresholds. To determine the upper threshold $t_{up}$, we run an experiment on randomly inserting ten corrupted labels on CIFAR-10 and roughly adjust the threshold so that the samples with corrupted labels can be automatically filtered out. The lower threshold $t_{low}$ is determined via a grid search. Algorithm~\ref{alg:gradnorm-sel} shows the process of the sample grouping using GradNorm.

\begin{algorithm}[b]
    \caption{Grouping samples using GradNorm}
    \label{alg:gradnorm-sel}
    \begin{algorithmic}[1]
    \REQUIRE Training set $\mathcal X$, the number of epochs $n$, the lower threshold $t_{low}$, the upper threshold $t_{up}$, learning rate $\gamma$, learning rate scheduler $s$
    \FOR{epoch$\in\{1,2,\dots, n\}$}
        \STATE For all $\pmb x_j$ in $\mathcal X$, compute $GradNorm(\pmb\theta, \{\pmb x_j\})$
        \STATE $\mu\leftarrow \dfrac{1}{|\mathcal X|}\sum\limits_{j=1}^{|\mathcal X|} GradNorm(\pmb\theta, \{\pmb x_j\})$
        \STATE $\mathcal S\leftarrow\{\pmb x\in \mathcal X|t_{low}\mu<GradNorm(\pmb\theta, \{\pmb x_j\})<t_{up}\mu\}$
        \STATE $\gamma\leftarrow \dfrac{|\mathcal S|}{|\mathcal X|}\gamma$
        \STATE Train the model on $\mathcal S$
        \STATE $\gamma\leftarrow \dfrac{|\mathcal X|}{|\mathcal S|}\gamma$
        \STATE $\gamma\leftarrow s(\gamma)$
    \ENDFOR
    \end{algorithmic}
\end{algorithm}

\subsection{Frequency-based Coreset Construction}

To select a coreset of a training set, GradNorms are first computed in the full set training process using the target model architecture, e.g., a ResNet-18. Then, the statistics are computed, i.e., the frequency of each data sample being selected by the aforementioned GradNorm thresholding strategy. 
Excluding the last overfitting epochs, we pick samples from subsets, being selected for more than $N$ epochs. In the case that the selected samples exceed the pre-defined budget, 
our method instead takes a fixed amount of data from the selected samples randomly. Algorithm~\ref{alg:ours-coreset} demonstrates our coreset construction method.

\begin{algorithm}[t]
    \caption{Frequency-based Coreset Construction}
    \label{alg:ours-coreset}
    \begin{algorithmic}[1]
    \REQUIRE Training set $\mathcal X$ for $n$ epochs, the number of epochs $n$, the lower threshold $t_{low}$, the upper threshold $t_{up}$, minimum frequency threshold $N$, the size of target coreset $s$
    \STATE Train the source model on $\mathcal X$, and record $GradNorm_t(\pmb\theta, \{\pmb x_j\})$ for every epoch $t$ and every data point $\pmb x_j\in \mathcal X$
    \STATE Initialize counting vector $\pmb c=[c_1,c_2,\dots, c_{|\mathcal X|}]\leftarrow[0,0,\dots, 0]$
    \FOR{$t\in\{1,2,\dots, n\}$}
        \STATE $\mu_t\leftarrow \dfrac{1}{|\mathcal X|}\sum\limits_{j=1}^{|\mathcal X|}GradNorm_t(\pmb\theta, \{\pmb x_j\})$
        \FOR{$j\in\{1,2,\dots, |\mathcal X|\}$}
            \IF{$t_{low}\mu_t<GradNorm_t(\pmb\theta, \{\pmb x_j\})<t_{up}\mu_t$}
                \STATE $c_j\leftarrow c_j + 1$
            \ENDIF
        \ENDFOR
    \ENDFOR
    \STATE $\mathcal S\leftarrow\{j\in\{1,2,\dots, |\mathcal{X}|\}| c_j\ge N\}$
    \IF{$|\mathcal S|>s$}
        \STATE $\mathcal S\leftarrow {\rm UniformSampling}(\mathcal S, s)$
    \ENDIF
    \STATE Train the target model on $\mathcal S$
    \end{algorithmic}
\end{algorithm}

\subsection{Learning Rate Adjustment}

Although the data that are not selected by the proposed selection method, either Data SI or GradNorm, are usually contributing little to the training (with small losses), 
we found that the learning rate should be adjusted to compensate for the reduced number of data samples so that the update steps are kept efficient as in training with the full set. Suppose that $\alpha$ portion of the full training set is selected, and the gradients of the remaining $(1-\alpha)$ portion are close to zeros. Naively aligning the batch updates on each parameter between the full set and our selected subset, we adjust the learning rate to be $\gamma'=\alpha\gamma$, where $\gamma$ is the original learning rate and $\gamma'$ is the learning rate to use. 

\subsection{Selection Overhead and Acceleration}
The computation cost of the proposed data subset selection using Data SI and GradNorm can be high. To address this issue, an approximation method utilizing the prediction scores $\pmb y$ can be applied. The loss is a function of the output predictions scores $\mathcal L(\pmb\theta, \mathcal X)=\mathcal L(\pmb y(\pmb\theta, \mathcal X))$, and we estimate $\Delta\pmb y$ with $\beta(\hat{\pmb y}-\pmb y)$ in Eq.~\ref{eq:approx}, where $\beta$ is a normalization factor and $\hat{\pmb y}$ is the ground truth label. Similarly, GradNorm is approximated via the prediction gradient norm.
\begin{equation}
    \label{eq:approx}
    \int_{\pmb\theta}^{\pmb\theta+\Delta\pmb\theta}\left(\dfrac{\partial \mathcal L}{\partial \pmb\theta}\right)^T{\mathrm d} \pmb\theta=\int_{\pmb y}^{\pmb y+\Delta\pmb y}\left(\dfrac{\partial \mathcal L}{\partial \pmb y}\right)^T{\mathrm d} \pmb y\approx \left(\dfrac{\partial \mathcal L}{\partial \pmb y}\right)^T\Delta\pmb y
\end{equation}

\section{Experiments and Results}

\textbf{Datasets.} A variety of real-world datasets are employed for the data redundancy investigation and following data selection experiments, i.e., CIFAR-10\cite{krizhevsky2009learning}, MIT Indoor Scenes (Indoor)\cite{quattoni2009recognizing}, Kvasir Capsule\cite{smedsrud2021kvasir} and ISIC 2019\cite{codella2018skin,combalia2019bcn20000,tschandl2018ham10000}. \textbf{CIFAR-10} is a popular set of 50,000 images in 10 classes with balanced image numbers for each class. \textbf{Indoor} is a natural image dataset with regular image sizes, containing 15620 images from 67 classes. \textbf{Kvasir Capsule} is a clinical video capsule endoscopy (VCE) dataset consisting of 47,238 labeled frames from 14 different classes. The \textbf{ISIC 2019} dataset is a dataset of real-world dermoscopic images containing 25,331 images from 9 classes. 

Images in the experiments are uniformly resized or cropped from the original large images (Indoor) as 224x224 image patches. For CIFAR-10, data augmentations are applied as the same as~\cite{he2016deep}. For Indoor, random cropping is conducted, together with random horizontal flips and normalization. For Kvasir Capsule, the only data pre-processing is normalization. As for ISIC 2019, we adopt the augmentation method used in CASS\cite{singh2022cass}. 

\textbf{Implementation details.} For all the experiments, we initialize image classification models (pre-trained on ImageNet) using weights in Torchvision for faster and more stable convergence, which also better satisfies first-order Taylor approximation conditions. The experiments are run on the SGD optimizer for 32 epochs, with a learning rate of 0.01, a multi-step learning rate scheduler that divides the learning rate by ten at the 16th and 24th epochs, the momentum of 0.9, weight decay of 0.0001, and early stopping with a tolerance of 5 epochs. All experiments are conducted on a single A100 on a DGX cluster. For the proposed adaptive data selection methods, we set the number of clusters $k=10$, upper and lower bounds $t_{up}=40$ times full set and $t_{lower}=0.1$ times full set, respectively. As for our coreset selection method, the frequency threshold is set to $N=4$ epochs.

\textbf{Compared methods.} Uncertainty-based active learning methods can be strong baselines for data subset selection\cite{park2022active}, and hence we compare Data SI and GradNorm with \textbf{Uniform} random sampling, the state-of-the-art adaptive data subset selection method \textbf{GradMatch}~\cite{killamsetty2021grad}, and two uncertainty-based active learning methods of selecting samples with least margin (\textbf{Margin})\cite{citovsky2021batch,netzer2011reading,nguyen2022measure,roth2006margin,zhan2022comparative} and least confidence (\textbf{Confidence})\cite{nguyen2022measure,wang2014new,zhan2022comparative}. We utilize the official GradMatch implementation (with GradMatchPB) and apply Torchvision pre-trained weights to the models. We keep most of the parameter settings as the default ones, except for adjusting the data selection ratio and frequency (once per epoch) according to our experiment. The train-validation split for GradMatch remains 9:1 as the original. For all the other compared methods, the train-validation split is 7:2 for datasets with an official test set split, and the train-validation-test split of 7:2:1 for those without official data splits.

\subsection{Observing Data SI Features}

\begin{figure}[t]
    \centering
    \includegraphics[width=0.95\textwidth]{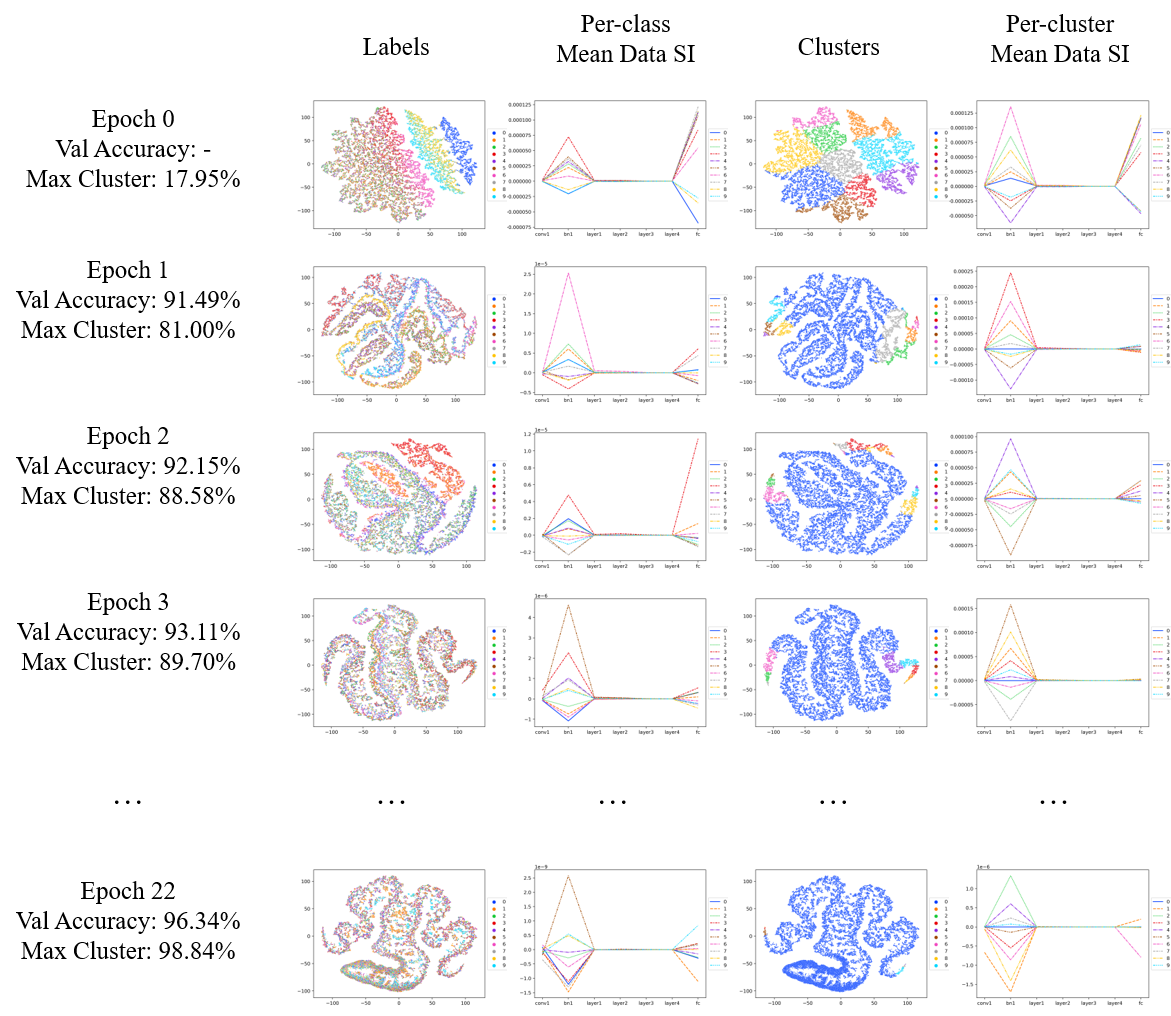}
    \caption{Per-epoch Data SI values. Columns from left to right indicate: the epoch information (epoch number, validation accuracy, and the percentage of the maximum cluster), the t-SNE Data SI scatter plot for each class, the per-class mean Data SI on each model (ResNet-18) layer, the t-SNE Data SI scatter plot clustered by KMeans, the per-cluster mean Data SI on each model layer.}
    \label{fig:per_epoch_si}
\end{figure}

As an example, the layerwise Data SI features are investigated and illustrated in training a ResNet-18 model on the CIFAR-10 in Fig.~\ref{fig:per_epoch_si}. Data SI features are shown with the t-SNE embedding, and the per-class and per-cluster mean Data SI on each model per layer are shown as well in the third and fifth columns. As seen in the `Clusters' column, the largest `blue' cluster grows larger along the training epoch and takes up the majority while having mean Data SI close to the axis (as in the far right column). Therefore, the majority of samples in most epochs are rather redundant (left-out ones) from the model training perspective, and the amount of them generally increases across the training process. Moreover, mean Data SI in BatchNorms and the last layer are observed to be more significant than the other layers, which may imply that these layers should be adapted more in the transfer learning setting.

\subsection{Results on Adaptive Data Selection}

\begin{figure}[tb]
    \centering
    \includegraphics[width=0.6\textwidth]{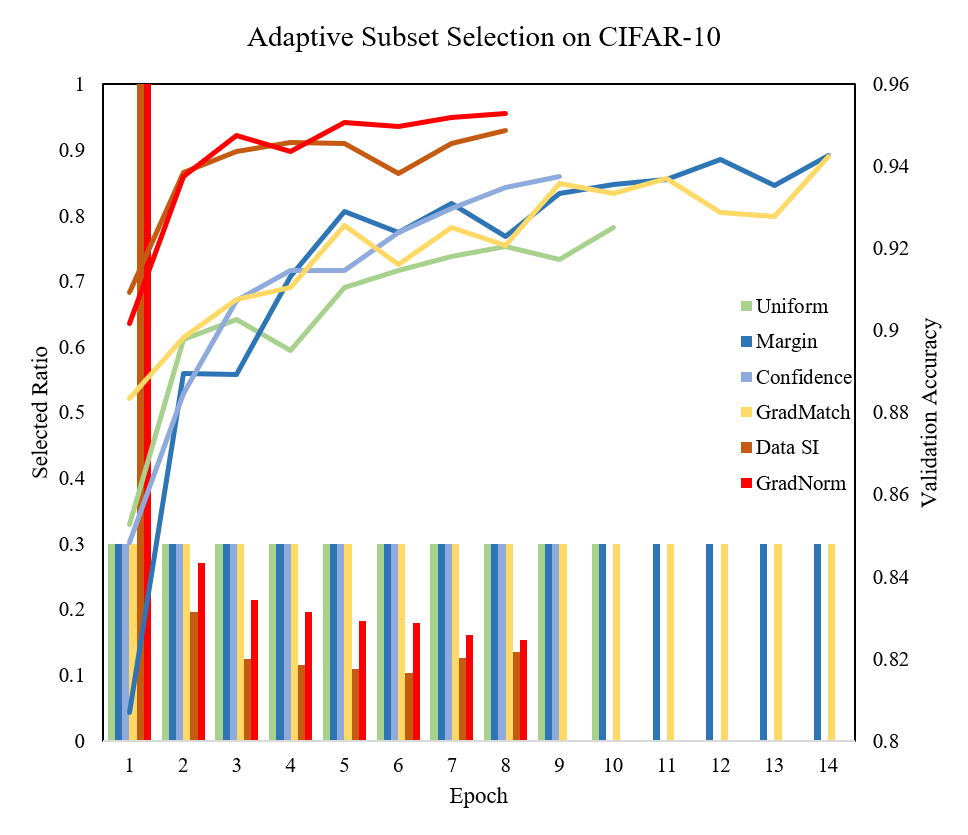}
    \caption{Adaptive selection process using the proposed selection methods. The bars indicate the selection ratios, and the lines indicate the classification accuracies (validation) for each epoch.}
    \label{fig:sel_process}
\end{figure}

Fig.~\ref{fig:sel_process} shows the adaptive selection experiment conducted on CIFAR-10. The selected amount of data (as ratios to the full set) and classification accuracy for each compared method are illustrated across training epochs.  Both Data SI and GradNorm-based data selection demonstrate a significant decrease in the data portion used in each epoch, while the other methods select a fixed portion of data randomly. Similar setting to warm-start using full set\cite{killamsetty2021grad}, our methods also apply a soft start strategy by first using a larger amount of data and then decreasing the amount gradually. The proposed methods also achieve better classification accuracy compared to other priors.

\begin{table}[tb]
    \centering
    \caption{Adaptive Subset Selection Test Performance on ResNet-18. Acc: test accuracy; BA: test balanced multi-class accuracy; A/E($\times$): average data portion ($\times$ the full training set) per epoch; Tot($\times$): total data amount ($\times$ full set) used in parameter update across all epochs. Balanced multi-class accuracy is taken due to imbalanced distribution in the datasets except for CIFAR-10.}
    \resizebox{\textwidth}{!}{
        \begin{tabular}{crrr|rrr|rrr|rrr}
        \toprule
             & \multicolumn{3}{c}{CIFAR-10} & \multicolumn{3}{c}{Indoor} & \multicolumn{3}{c}{Kvasir Capsule} & \multicolumn{3}{c}{ISIC 2019} \\
             & Acc$\uparrow$ & A/E$\downarrow$  & Tot $\downarrow$ & BA$\uparrow$ & A/E$\downarrow$ & Tot$\downarrow$ & BA$\uparrow$ & A/E$\downarrow$ & Tot$\downarrow$ & BA$\uparrow$ & A/E$\downarrow$ & Tot$\downarrow$ \\
        \midrule
            Full Set & .959 & 1.00 & 30.00 & .697 & 1.00 & 13.00 & .969 & 1.00 & 22.00 & .578 & 1.00 & 10.00 \\
        \midrule
           Uniform & .927 & .30 & 4.50 & \textbf{.713} & .65 & 20.80 & .927 & .40 & 5.20 & .535 & .60 & 6.00\\
           GradMatch & .943 & .30 & 5.70 & .683 & .65 & 19.00 & .972 & .40 & 7.20 & .660 & .60 & 11.40 \\
           Margin & \textbf{.959} & .30 & 8.70 & .645 & .65 & 8.45 & .970 & .40 & 5.20 & \textbf{.702} & .60 & 13.80 \\
           Confidence & .958 & .30 & 8.40 & .656 & .65 & 10.40 & \textbf{.977} & .40 & 10.00 & .565 & .60 & 5.40\\
        \midrule
           Data SI & .948 & \textbf{.19} & 2.43 & \textbf{.722} & \textbf{.59} & 16.60 & .937 & .38 & 3.82 & \textbf{.654} & \textbf{.54} & 9.13\\
           GradNorm & \textbf{.954} & .22 & 3.30 & .701 & .71 & 12.00 & \textbf{.940} & \textbf{.22} & 3.03 & .625 & .55 & 9.38\\
        \bottomrule
        \end{tabular}
    }
    \label{tab:exp-2}
\end{table}

\textbf{Performance on different datasets.} We apply Data SI and GradNorm-based selection strategies on two natural image datasets, i.e., CIFAR-10 and Indoor, and two medical image classification datasets, i.e., Kvasir Capsule and ISIC 2019, with larger, noisy and more imbalanced image data and labels.

\begin{figure}[t]
    \centering
    \subfloat[CIFAR-10]{\includegraphics[width=0.42\textwidth]{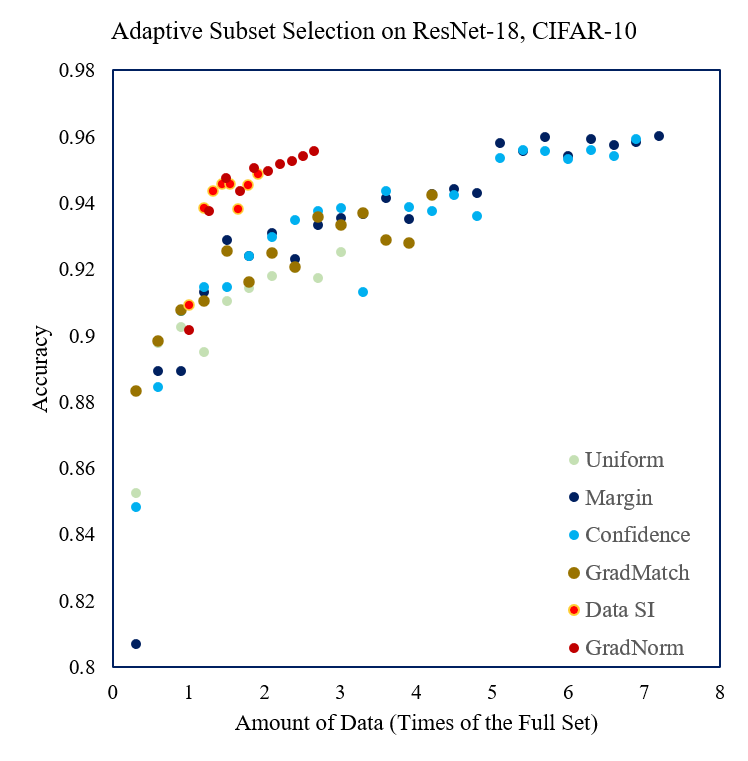}}
    \subfloat[Indoor]{\includegraphics[width=0.42\textwidth]{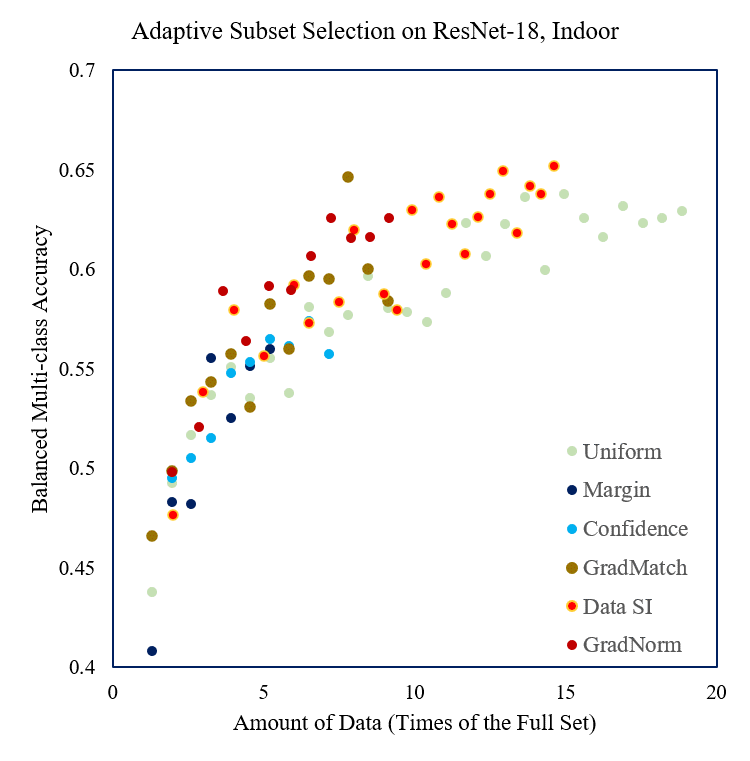}}   \\ 
    \subfloat[Kvasir Capsule]{\includegraphics[width=0.42\textwidth]{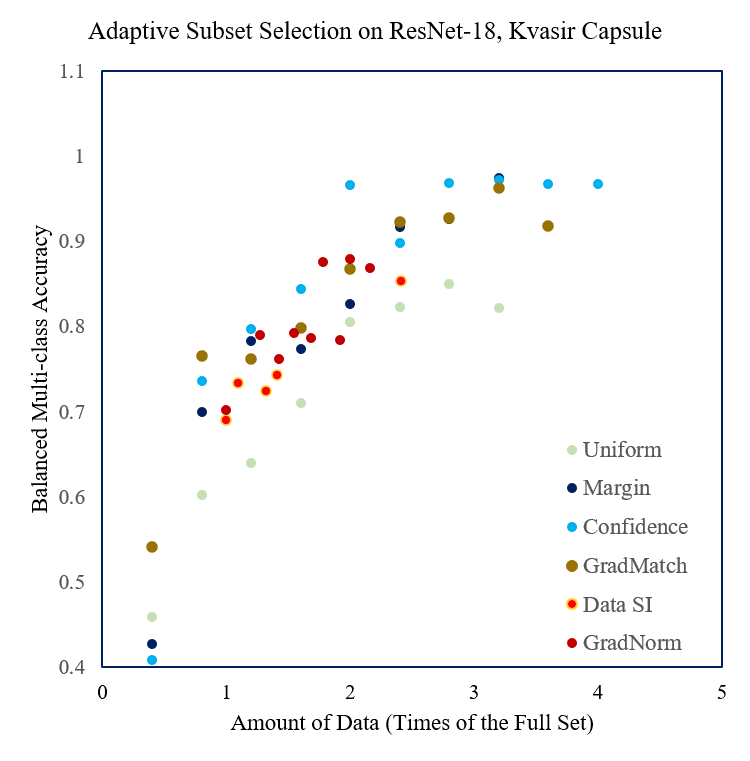}}
    \subfloat[ISIC 2019]{\includegraphics[width=0.42\textwidth]{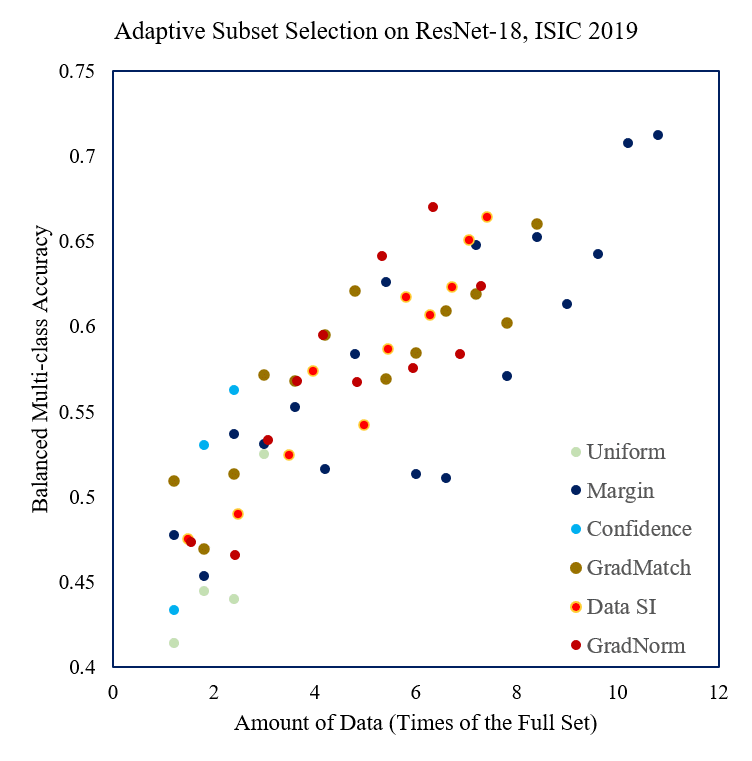}}
    \caption{Validation accuracy and total amount of data used for each selection method. The upper left indicates better performance. Data SI and GradNorm are marked with orange and red.}
    \label{fig:tradeoff-auc}
\end{figure}
The final balanced multi-class accuracies on the test set are reported in Table~\ref{tab:exp-2}. For Indoor, Data SI significantly outperforms the other methods.  Our methods achieve comparable test accuracy on CIFAR-10 and ISIC 2019 while less amount of data is used. On CIFAR-10, our methods used only approximately 3 times of the total data amount for the training, which is far smaller than all the other methods, while maintaining equivalent accuracy as using the full set. It often requires a balance of the amount of data used for the training and the output classification accuracy. The ratios (e.g., .30 on CIFAR-10) for the compared methods are set to larger tenths based on the A/E generated by the proposed methods.

To fully illustrate the relation between the selected amount of data and prediction accuracy, the validation accuracies and total amount of data used are visualized in AUC-like plots Fig.~\ref{fig:tradeoff-auc}. In these figures, points closer to the upper left corner demonstrate better performance as they achieve higher accuracy with fewer data. On CIFAR-10, Indoor, and ISIC 2019, our methods (marked with orange and red) are overall closer to the upper left corner. On ISIC 2019 and Kvasir Capsule, the methods except for uniformly random selection are close to each other, i.e. non-uniform sampling strategy can perform better data amount-accuracy trade-off than the uniform one. 

\textbf{Performance on different network architectures.} Table~\ref{tab:diff-arch} shows how the proposed adaptive data selection performs on a variety of network architectures from convolutional neural networks to the transformer model, i.e., ResNet-18, ResNet-50, EfficientNetV2-S and ViT-B-16 trained on CIFAR-10. Both GradNorm and Data SI are applicable to various model architectures and can usually achieve superior performance in comparison to priors.

\begin{table}[tb]
    \centering
    \caption{The classification accuracy of different adaptive subset selection methods across various model architectures on CIFAR-10}
    \begin{tabular}{ccccc}
    \toprule
        ACC & ResNet-18 & ResNet-50 & EfficientNetV2-S & ViT-B-16 \\
    \midrule
       Full Set & .959 & .964 & .979 & .970 \\
    \midrule
        Uniform 30\% & .927 & .948 & .963 & .945  \\
        Margin 30\% & \textbf{.959} & .959 & .974 & .969\\
        Confidence 30\% & .958 & .963 & .974 & \textbf{.979}\\
        \midrule
        Data SI 13\%-20\% & .948 & .956 & \textbf{.976} & .978 \\
        GradNorm 12\%-22\% & .954 & \textbf{.965} & .975 & \textbf{.979} \\
    \bottomrule
    \end{tabular}
    \label{tab:diff-arch}
\end{table}

\subsection{Results on Coreset Selection}

\begin{table}[tb]
    \centering
    \caption{Test balanced multi-class accuracies for coresets selected on ResNet-18 and retrained on ResNet-18. $^l$ indicates that the method is trained following the same learning rate adjustment as ours.}
    \begin{tabular}{lccccccccc}
    \toprule
         & \multicolumn{2}{c}{CIFAR-10} & \multicolumn{2}{c}{Indoor} & \multicolumn{2}{c}{Kvasir Capsule} & \multicolumn{2}{c}{ISIC 2019}  \\
         & Acc$\uparrow$ & Sz($\times$)$\downarrow$ & BA$\uparrow$ & Sz($\times$)$\downarrow$ & BA$\uparrow$ & Sz($\times$)$\downarrow$ & BA$\uparrow$ & Sz($\times$)$\downarrow$\\
    \midrule
       Uniform & .927 & .30 & \textbf{.609} & .30 & .892 & .30 & .400 & .30 \\
       Uniform$^l$ & .933 & .30 & .558 & .30 & .794 & .30 & .435 & .30 \\
       CDAL & .918 & .30 & .564 & .30 & .889 & .30 & .433 & .30 \\
       CDAL$^l$ & .929 & .30 & .464 & .30 & .795 & .30 & .458 & .30  \\
       Uncertainty & .902 & .30 & .557 & .30 & .458 & .30 & .322 & .30\\
       Uncertainty$^l$ & .898 & .30 & .484 & .30 & .408 & .30 & .372 & .30\\
       DeepFool & .926 & .30 & .553 & .30 & .661 & .30 & .419 & .30 \\
       DeepFool$^l$ & .925 & .30 & .461 & .30 & .595 & .30 & .392 & .30 \\
    \midrule
       Ours & \textbf{.945} & .24 & .570 & .30 & \textbf{.932} & .18 & \textbf{.493} & .30 \\
    \bottomrule
    \end{tabular}
    \label{tab:exp4-2}
\end{table}

 We follow the experimental setting of DeepCore\cite{guo2022deepcore} and compare the performance of models trained with 30\% randomly sampled data and three commonly used coreset approaches, including CDAL\cite{agarwal2020contextual}, Least Confidence, and DeepFool\cite{ducoffe2018adversarial}. The retraining results are provided in Table~\ref{tab:exp4-2}. The learning rate is adjusted as before since the contributing data will result in larger steps in optimization. The performance of the baseline methods with the same learning rate adjustment is also shown for a fair comparison. Our learning rate adjustment has little or no improvement on the performance of the baselines, and our selected (<=30\%) coreset consistently outperforms the 30\% coreset in the baselines on ResNet-18. Uniformly random sampling outperforms the other methods on Indoor, which may result from the lack of samples per class and the biased selected coreset.

\subsection{Other Experiment Results}
\begin{table}[tb]
    \centering
    \caption{Ablation study on the frequency threshold for coreset construction. The experiment is conducted on ISIC 2019 datasets, ResNet-18, and the coreset selected are all 30\% of the original.}
    \begin{tabular}{cccc}
    \toprule
        Frequency Threshold & 0 (Uniformly Random) & 4 & 8 \\
    \midrule
        Balanced Multi-class Accuracy &  40.04\% & \textbf{49.28\%} & 31.27\% \\
    \bottomrule
    \end{tabular}
    \label{tab:abl-coreset-freq}
\end{table}
\begin{table}[tb]
    \centering
    \caption{Ablation study on learning rate adjustment. The experiment is conducted on dataset CIFAR-10, ResNet-18.}
    \begin{tabular}{cccc}
    \toprule
        \% & Adjust & No adjust & Gain by adjustment\\
    \midrule
        GradNorm & 95.44 & 92.74 & +2.70\\
        Data SI & 94.80 & 91.92 & +2.88\\
    \bottomrule
    \end{tabular}
    \label{tab:abl-lr-discount-si}
\end{table}
\begin{table}[tb]
    \centering
    \caption{Approximation Methods, ResNet-18, CIFAR-10, Balanced Multi-class Accuracies}
    \begin{tabular}{ccccc}
    \toprule
       \%  & CIFAR-10 & Indoor & Kvasir Capsule & ISIC 2019 \\
    \midrule
       Full Set & 95.90 & 69.67 & 96.85 & 57.79\\
   \midrule
       Data SI & 94.80 & 72.24 & 93.72 & 65.44\\
       GradNorm & 95.44 & 70.14 & 93.99 & 62.53\\
        Data SI Approx & 95.36 & 70.56 & 96.13 & 54.70\\
        GradNorm Approx & 95.30 & 71.19 & 86.63 & 69.27\\
    \bottomrule
    \end{tabular}
    \label{tab:approx}
\end{table}
\textbf{Frequency threshold for coreset construction.} We set the frequency threshold to be 4, which is balanced between including as many contributing samples and maximizing average sample contributions. For fair coreset size comparison, we demonstrate such a trade-off on ISIC 2019 in Table~\ref{tab:abl-coreset-freq}, where the chosen threshold achieves better performance than other values.

\textbf{Learning rate Adjustment.} During the experiments, we found that learning rate adjustment can usually help in training. As an example, the test performance of CIFAR-10 for GradNorm with and without the learning rate adjustment is shown in Table~\ref{tab:abl-lr-discount-si}, where learning rate adjustment shows 2.70\% more accuracy.

\textbf{The number of clusters.} Table~\ref{tab:abl_k} presents the performance of our adaptive data selection approach, with varying numbers of clusters $k$ on Data SI features. Generally, smaller $k$ will induce a less aggressive selection strategy and select fewer samples for training, leading to a larger performance drop.

\begin{table}[t]
    \centering
    \begin{tabular}{cccccc}
    \toprule
        $k$ & 3 & 5 & 10 & 15 & 30 \\
    \midrule
        Acc $\uparrow$ & .931 & .932 & .948 & .952 & .955 \\
        Tot ($\times$) $\downarrow$ & 1.89 & 1.84 & 2.43 & 3.41 & 3.11 \\
        A/E ($\times$) $\downarrow$ & .27 & .26 & .19 & .17 & .24 \\
    \bottomrule
    \end{tabular}
    \caption{The effect of different numbers of clusters in adaptive data selection using Data SI features. Acc: test accuracy; Tot: the total amount of data (times full set) used in training; A/E: the average amount of data (times full set) used in training}
    \label{tab:abl_k}
\end{table}
 
\textbf{SI Approximation.} We can approximate Data SI by batch forwarding and back-propagation to the output scores, reducing the valuation and selection overhead to <30 seconds for 50,000 224x224 RGB images on ResNet-18 on a single A100. 

The testing multi-class accuracies using the approximation of the GradNorm are reported in Table~\ref{tab:approx}, where the approximation methods work well in CIFAR-10 and Indoor while yielding unstable performance on Kvasir Capsule and ISIC 2019, which is optioned out in the experiments above.

\begin{figure}[b]
    \centering
    \subfloat[Some of the easy samples of Angiectasia]{\includegraphics[width=0.75\textwidth]{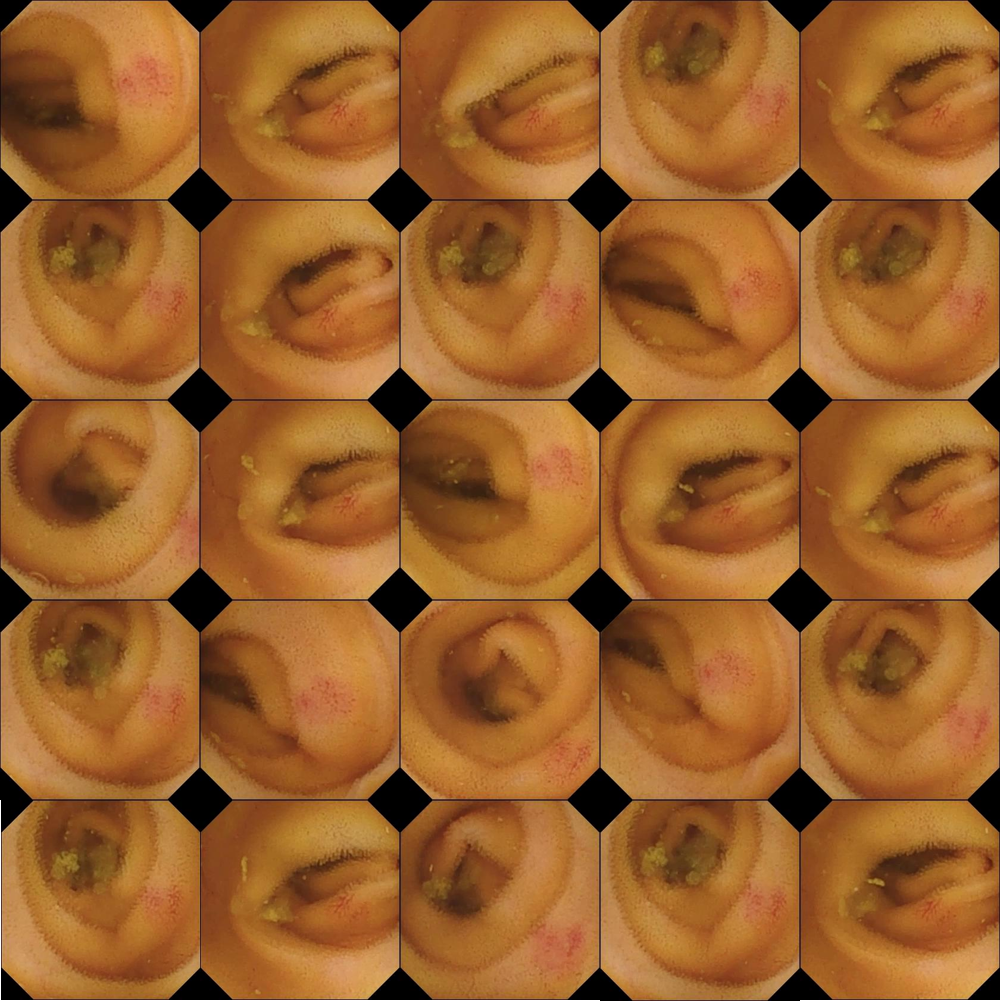}}
    
    \subfloat[Some of the hard samples of Angiectasia]{\includegraphics[width=0.75\textwidth]{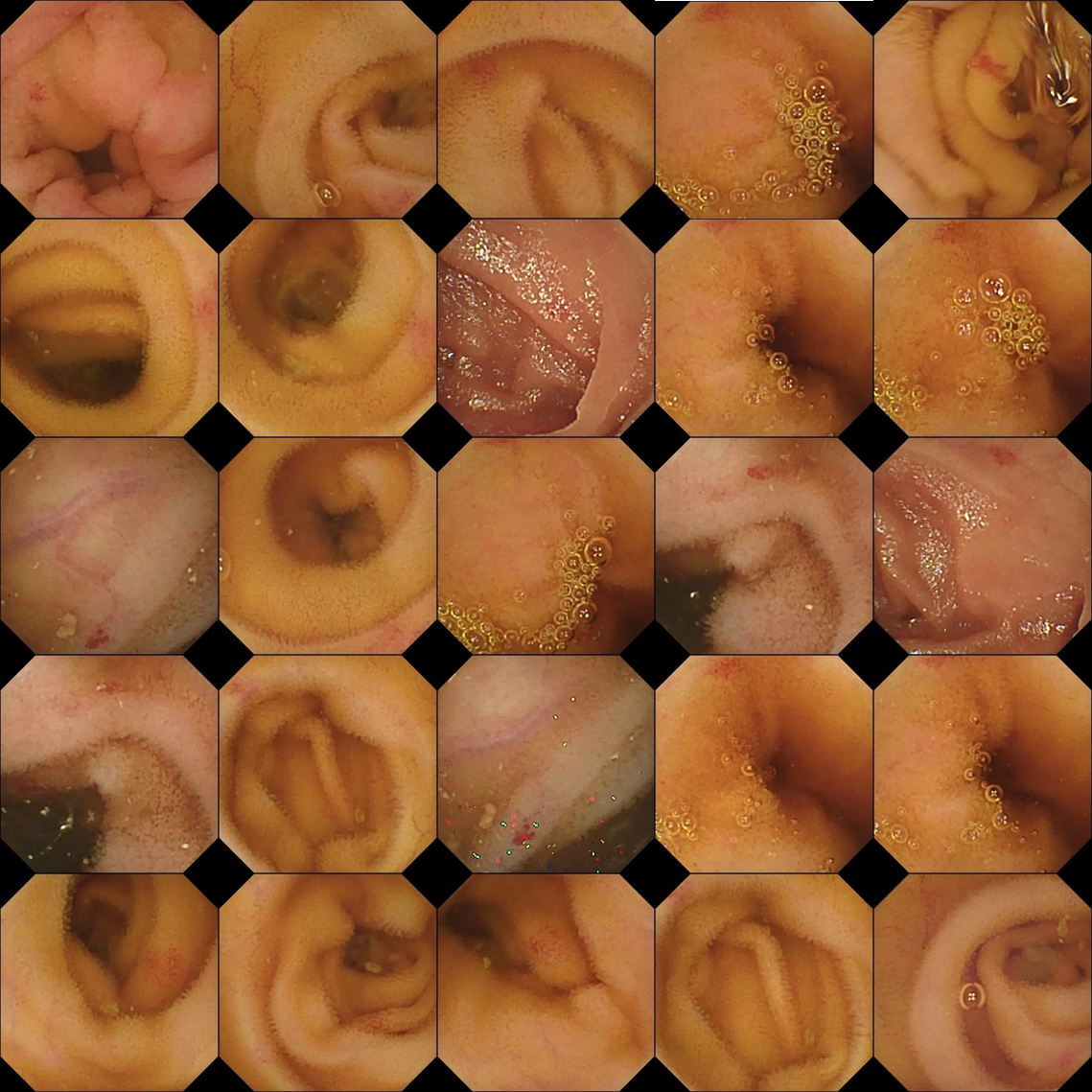}}
    \caption{Easy and hard samples for Angiectasia}
    \label{fig:angiectasia}
\end{figure}

\begin{figure}[b]
    \centering
    \subfloat[Some of the easy samples of Blood - Fresh]{\includegraphics[width=0.75\textwidth]{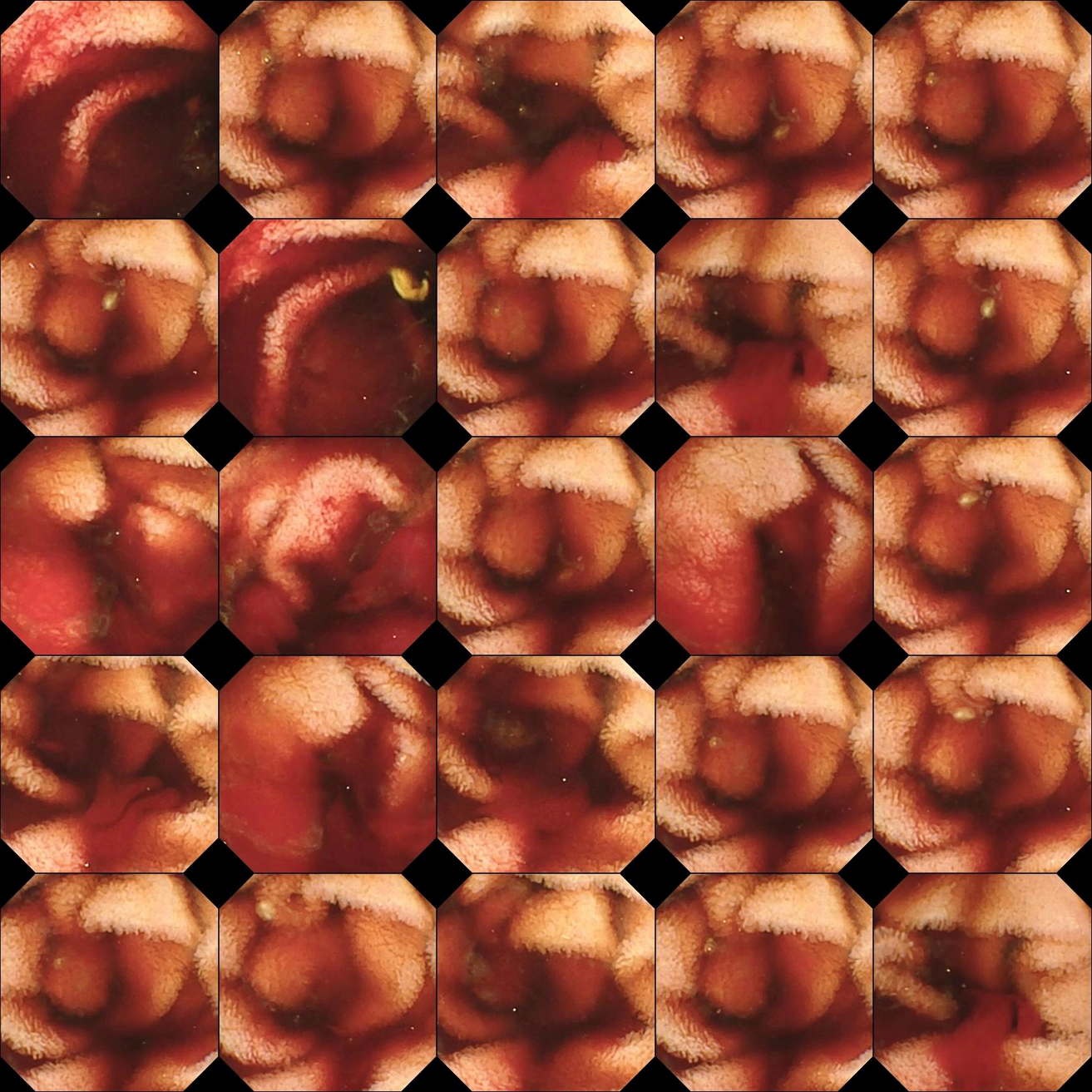}}
    
    \subfloat[The hard samples of Blood - Fresh]{\includegraphics[width=0.6\textwidth]{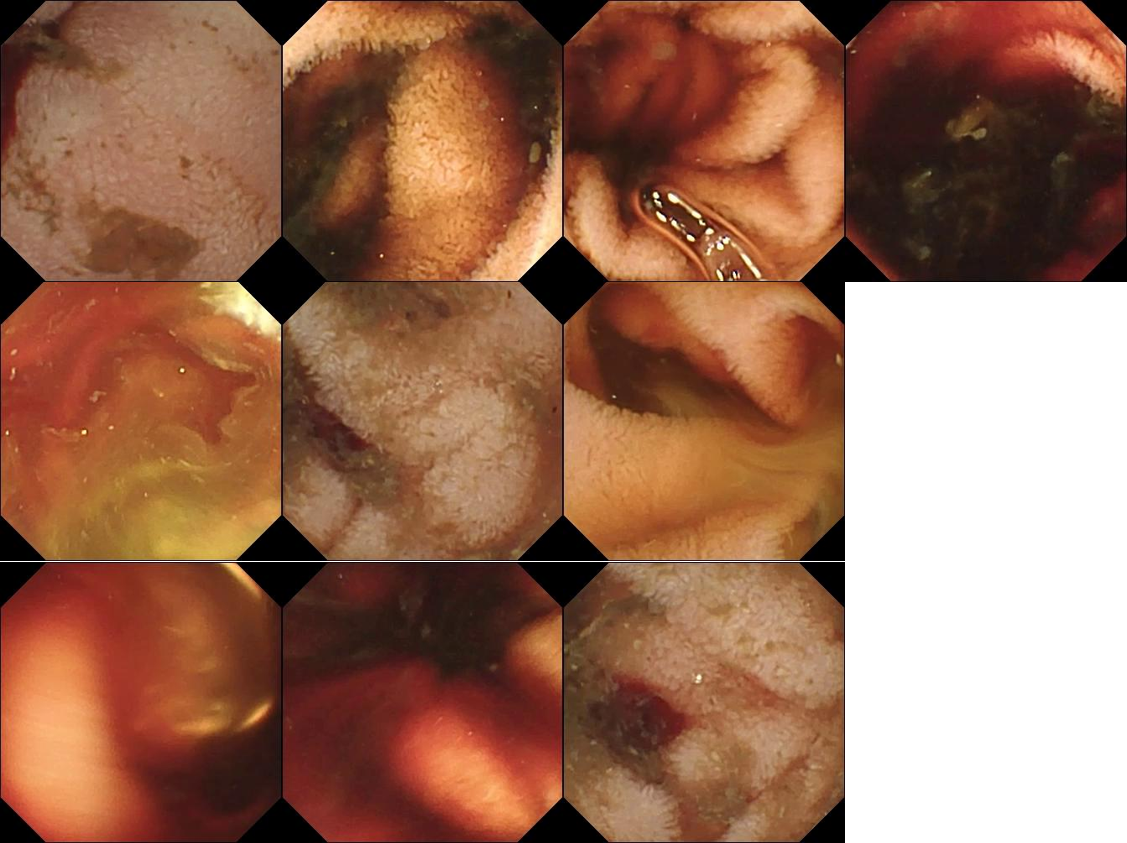}}
    \caption{Easy and hard samples for Blood - Fresh}
    \label{fig:blood_fresh}
\end{figure}

\begin{figure}[b]
    \centering
    \subfloat[Some of the easy samples of Foreign Body]{\includegraphics[width=0.75\textwidth]{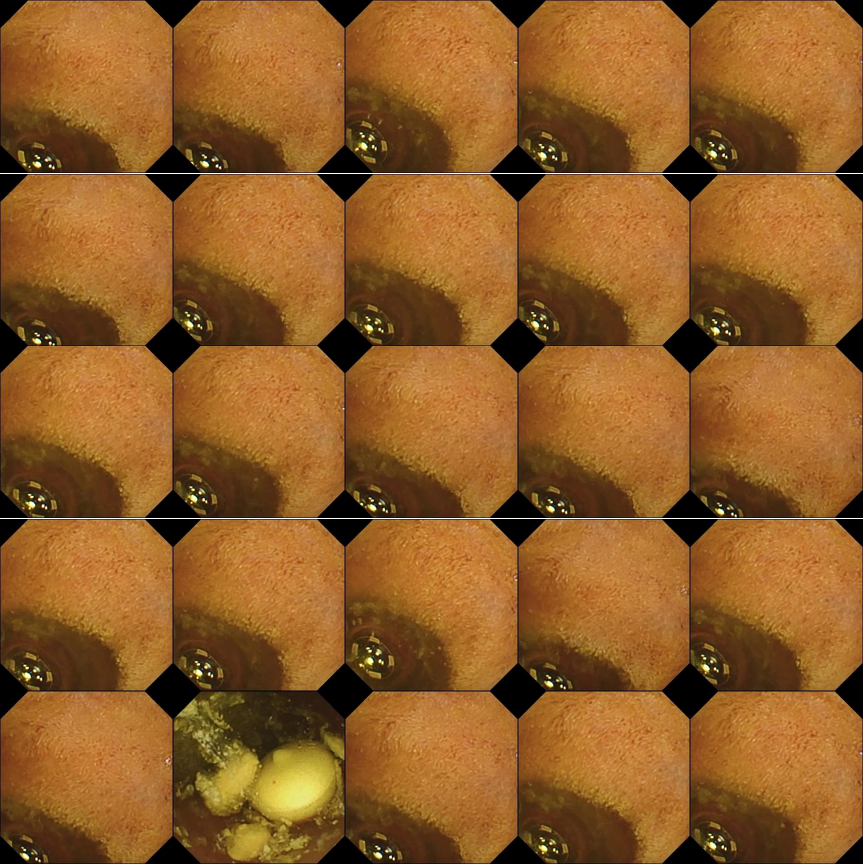}}
    
    \subfloat[Some of the hard samples of Foreign Body]{\includegraphics[width=0.75\textwidth]{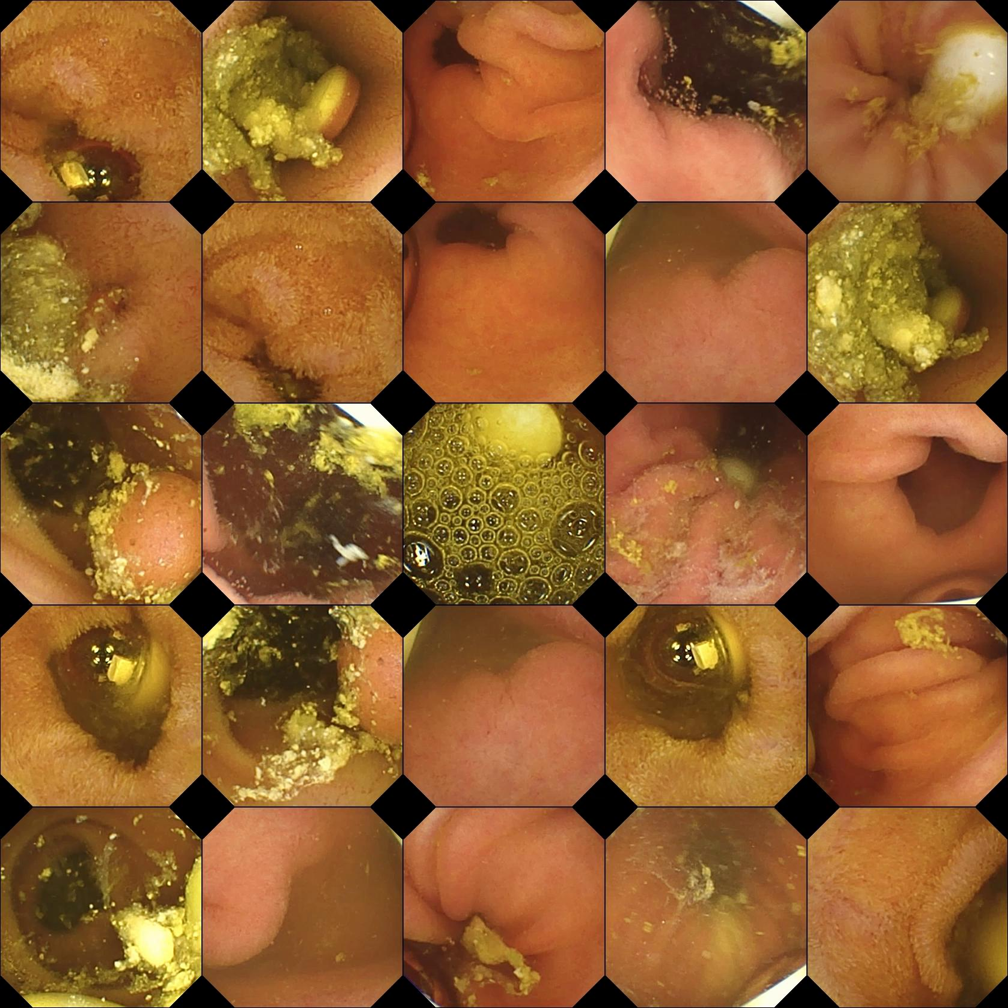}}
    \caption{Easy and hard samples for Foreign Body}
    \label{fig:foreign_body}
\end{figure}

\textbf{Easy and hard samples.} Similar to our coreset selection approach, the frequency of samples to be selected in each epoch is recorded. We study the samples in Kvasir Capsule that are least selected (for one epoch, called easy samples here) and most selected (for 23 epochs, called hard samples here). Some easy and hard samples of three classes are presented in Fig.~\ref{fig:angiectasia}, \ref{fig:blood_fresh}, and \ref{fig:foreign_body}. The easy samples are often observed to be duplicated and redundant, while the hard samples are more diverse.

\section{Limitations and Final Remarks}

This paper investigates layerwise gradient-based data valuation metrics Data SI and GradNorm in examining the data redundancy in real-world image datasets. Additionally, novel adaptive data selection methods are proposed, with automated and dynamic selected subsets for each training epoch. The valuation methods are derived from the allocation of loss change to data, and the selection strategies aim to greedily maximize the loss change in each epoch. The effectiveness of our valuation metrics and selection methods are demonstrated on cross-domain datasets and various model architectures. Online and offline data selection experiments show the efficiency of our coreset and the varied data redundancy in each dataset.

Our study is limited to the multi-class image classification task. Due to the different nature of image labels, we have not developed efficient adaptive subset selection methods on multi-label classification and more tasks with dense labels, e.g., semantic/instance segmentation. 
Besides, the presented work largely relies on the empirical results, though they show that our methods converge well on various model architectures and datasets. However, the methods derived from greedy optimization targets may not prove to be effective in model convergence from a theoretical perspective. 

We believe that the introduced two data valuation metrics and corresponding data selection strategies could be applied in many other tasks and scenarios. For example, our data valuation approach, which relies solely on gradients and activations, can be applied to the design of fairness and incentive mechanisms in federated learning. Furthermore, the constructed coreset demonstrates strong performance in retraining the model and may have potential applications in the construction of memory buffers for replay-based continual learning.
\bibliographystyle{abbrvnat}
\bibliography{reference.bib}

\end{document}